\documentclass[10pt,twocolumn,letterpaper]{article}

\usepackage{wacv}
\usepackage{times}
\usepackage{epsfig}
\usepackage{graphicx}
\usepackage{amsmath}
\usepackage{amssymb}
\usepackage{makecell}

\wacvfinalcopy 

\def\wacvPaperID{284} 

\ifwacvfinal\fi
\begin{document}

\title{How Much Chemistry Does a Deep Neural Network Need to Know to Make Accurate Predictions?}

\author{Garrett B. Goh \textsuperscript{1,*}, Charles Siegel \textsuperscript{2}, Abhinav Vishnu \textsuperscript{1}, Nathan Hodas \textsuperscript{2}, Nathan Baker \textsuperscript{1}\\
\and
\\
\textsuperscript{1} Advanced Computing, Mathematics \& Data Division, Pacific Northwest National Lab \\
\textsuperscript{2} Computing \& Analytics Division, Pacific Northwest National Lab \\
{\tt\small garrett.goh@pnnl.gov}
}

\maketitle
\ifwacvfinal\thispagestyle{empty}\fi

\begin{abstract}

The meteoric rise of deep learning models in computer vision research, having achieved human-level accuracy in image recognition tasks is firm evidence of the impact of representation learning of deep neural networks. In the chemistry domain, recent advances have also led to the development of similar CNN models, such as Chemception, that is trained to predict chemical properties using images of molecular drawings. In this work, we investigate the effects of systematically removing and adding localized domain-specific information to the image channels of the training data. By augmenting images with only 3 additional basic information, and without introducing any architectural changes, we demonstrate that an augmented Chemception (AugChemception) outperforms the original model in the prediction of toxicity, activity, and solvation free energy. Then, by altering the information content in the images, and examining the resulting model's performance, we also identify two distinct learning patterns in predicting toxicity/activity as compared to solvation free energy. These patterns suggest that Chemception is learning about its tasks in the manner that is consistent with established knowledge. Thus, our work demonstrates that advanced chemical knowledge is not a pre-requisite for deep learning models to accurately predict complex chemical properties. 

\end{abstract}

\section{Introduction}
\label{sec:intro}

The role of deep learning in transforming computer vision research is significant. Prior to deep learning, computer vision researchers reached a glass ceiling of 20-30\% top-5 error rate in image recognition tasks. Within 3 years of using deep neural network (DNN) models, human-level accuracy of under 5\% top-5 error rate was achieved, and DNN models have become the dominant algorithm in computer vision research~\cite{he2015,szegedy2015,szegedy2017}. In chemistry, DNN models were the winning entry in the Merck Kaggle challenge in 2012 and the NIH Tox21 challenge in 2014. Following this trend, several research groups have started using DNN models to predict numerous properties, including activity,~\cite{dahl2014,ma2015,ramsundar2015,unterthiner2014} toxicity,~\cite{mayr2016,xu2015} reactivity,~\cite{hughes2016,hughes2015}, solubility,~\cite{lusci2013}, ADMET properties,~\cite{kearnes2016} and various QM computed energies.~\cite{montavon2013,smith2017,schutt2017} Across the chemistry domain, DNN models are frequently performing as well as or better than previous state-of-the-art models based on traditional machine (ML) learning algorithms.~\cite{goh2017r,gawehn2016}

DNN models are capable of learning representations, which sets it apart from conventional ML algorithms used in chemistry. Representation learning is the process of transforming input data into a feature that can be effectively exploited to identify patterns from data. Historically, computer vision research invested significant effort in designing appropriate features using applied mathematics.~\cite{lowe1999} However, today, such expert-driven feature engineering research has been replaced by deep learning models and its representation learning ability, which is the primary reason why DNN models were able to exceed the glass ceilings of that field.~\cite{he2015,szegedy2017} In the chemistry context, the analogous process would be to use deep learning to examine chemical structures and to construct features similar to engineered chemical features, with minimal assistance from an expert chemist. This approach that leverages representation learning of deep neural networks, is a significant departure from the traditional research paradigm in chemistry.

We observe that recent trends in the literature have exhibited a subtle shift from relying heavily on engineered chemistry features (e.g. molecular fingerprints), to the development of engineered representations of a molecule. Specifically, the Aspuru-Guzik~\cite{duvenaud2015} and Pande groups~\cite{kearnes2016} have developed deep learning algorithms for analyzing molecular graphs, and using SMILES data (a text-based representation) was also found to be effective. Recent work by Goh et. al.~\cite{goh2017c1} on the Chemception CNN model has also demonstrated that even “unsophisticated” data like images of molecular drawings with minimal chemical information can be used to develop models that were on average equivalent to contemporary DNN models trained on molecular fingerprints.~\cite{goh2017c1} Despite these accomplishments, the lack of labeled data in chemistry (which requires costly experiments or simulations to generate labels) is an impediment to unlocking the full potential of representation learning of deep neural networks. 

\subsection{Contributions}

Building on the original Chemception work,~\cite{goh2017c1} our research provides two key contributions to the field: (i) an appropriate method for encoding domain-specific information in images to improve model performance, and (ii) the results of our experiments that illuminate the nature of how deep neural networks learn about chemical data.

First, we have developed a novel image augmentation technique that adds localized chemical information into the channels of the images used to train Chemception. The hypothesis behind this approach is that if relevant domain-specific information were to be provided to the neural network, it would not need to learn the representations for these basic features, but instead will be able to direct more of its learning capacity to develop more sophisticated representations and improve its accuracy for predicting complex chemical properties. Using these augmented chemical images, we show that it outperforms earlier Chemception models (without making any architecture changes), as well as most contemporary deep learning models for predicting a broad range of chemical properties, including toxicity, activity and solvation free energy. \textit{To the best of our knowledge, such an approach has yet to be reported in the chemistry deep learning literature, and manipulating image channel data with localized domain-specific information is also a relatively unexplored avenue in computer vision research.} 

Second, we performed comprehensive experiments on altering the information content encoded into image channels, and examined its impact on model accuracy. Our results demonstrate that a firm grasp of advanced chemical knowledge is not a pre-requisite for deep learning models to accurately predict chemical properties. Furthermore, based on Chemception's performance under various information content scenarios in predicting toxicity/activity as compared to solvation free energy, we identified that Chemception learns in a manner that appears to be consistent with established chemical knowledge. \textit{To the best of our knowledge, such an approach of altering information content in order to infer how/what a neural network learns has also not been reported in the literature.} 

\section{Methods}
\label{sec:background}

In this section, we provide details on the datasets used, data splitting and data preparation steps. Then, we document the training protocol and evaluation metrics used.

\subsection{Dataset Description}
Following an identical approach as the original Chemception paper,~\cite{goh2017c1} we obtained the Tox21, HIV, and FreeSolv datasets (Table~\ref{table:1}) from the MoleculeNet benchmark database,~\cite{wu2017} to evaluate the performance of Chemception for predicting toxicity, activity and solvation free energy respectively. These datasets comprise of a mix of large vs small datasets, physical vs non-physical properties and regression vs classification problems. In addition, for solvation free energy, we also included data from alchemical free energy calculations, which is used as a target comparison against existing physics-based models.

\begin{table}[!t] 
		\begin{center}
		\begin{tabular}{|c|c|c|c|}
				\hline
				Dataset & Property & Task & Size \\
				\hline\hline
				Tox21 & \makecell{Non-Physical \\(Toxicity)} & \makecell{Multi-task \\ classification} & 8014 \\
				\hline
				HIV & \makecell{Non-Physical \\(Activity)} & \makecell{Single-task \\ classification} & 41,193 \\
				\hline
				FreeSolv & \makecell{Physical \\(Solvation)} & \makecell{Single-task \\ regression} & 643\\
				\hline
		\end{tabular} 
		\end{center}
\caption{Characteristics of the 3 datasets used in model evaluation.}
\label{table:1}
\end{table}

\subsection{Data Preparation}
\begin{figure}[!htbp]
\centering
\includegraphics[scale=0.5]{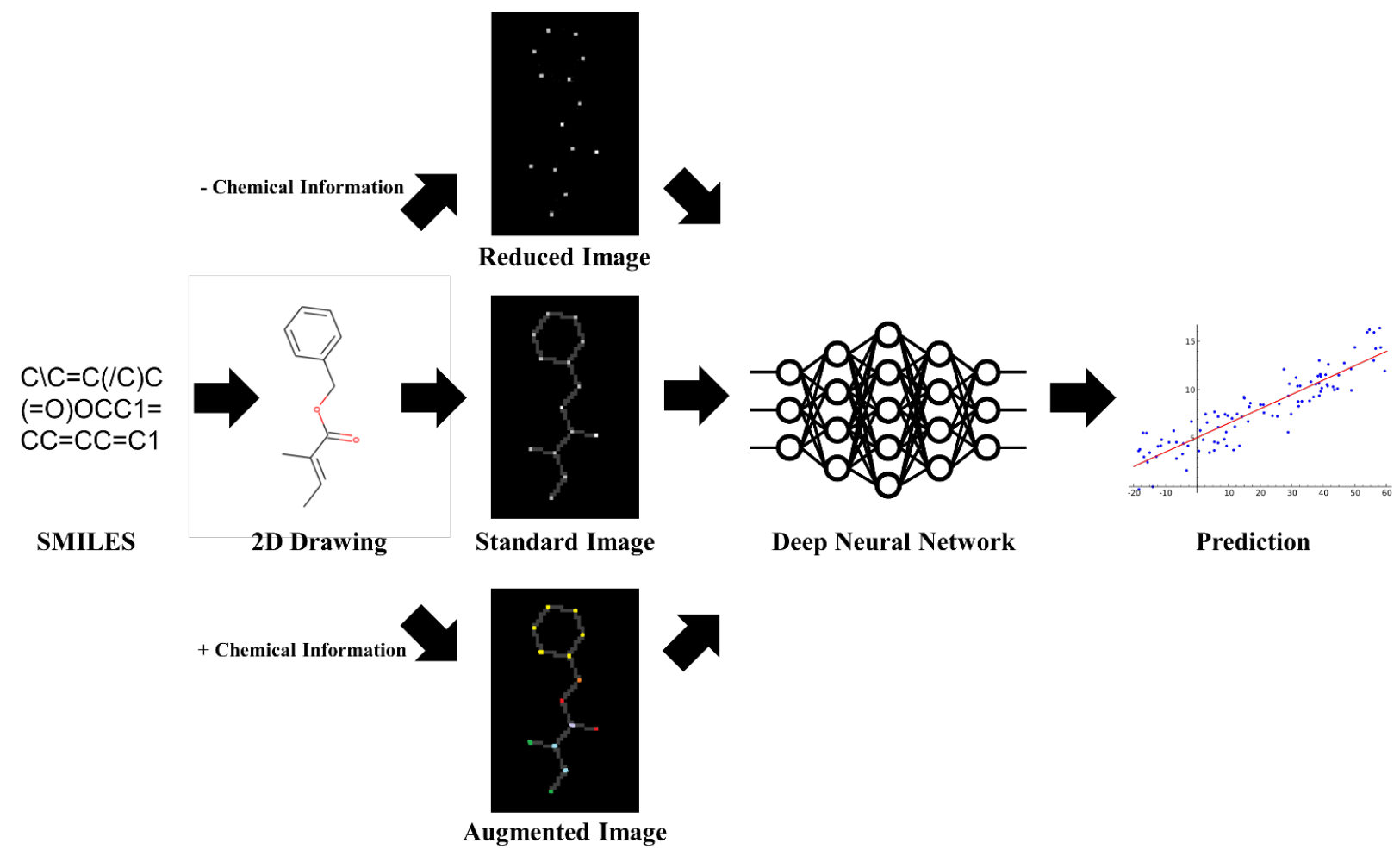}
\caption{\small After a SMILES to structure conversion, the 2D images are discretized and used to train a deep neural network. In this work, different image representations were used. Augmented images included localized atom and bond-specific chemical information, and reduced images had simplified information encoded into a single image channel.}
\label{fig:1}
\end{figure}

A summary of the end-to-end workflow of Chemception is illustrated in Figure~\ref{fig:1}. Briefly, we used open-source cheminformatics software, RDKit~\cite{landrum2016} to convert SMILES strings to their respective 2D molecular structures. The resulting coordinates of each molecule were then mapped onto a discretized image of 80 x 80 pixels that corresponds to 0.5 Å resolution per pixel, which was used to train Chemception. In this work, we varied the information content in the image channels to create several different augmented and reduced information image formats, and their details are outlined in the Results section.

\subsection{Data Splitting}
The dataset pre-processing steps are identical to that reported previously.~\cite{goh2017c1}  We used a 5-fold cross validation protocol for training and evaluated the performance and early stopping criterion of the model using the validation set. We also included the performance on a separate test set as an indicator of generalizability. Specifically, for the Tox21 and HIV dataset, 1/6th of the database was separated out to form the test set, and for the Freesolv dataset, 1/10th of the database was used to form the test set. The remaining 5/6th or 9/10th of the dataset was then used in the random 5-fold cross validation approach for training Chemception. 

For classification tasks (Tox21, HIV), we also over-sampled the minority class to address the class imbalance observed in the dataset, by appending additional data from the minority class by a factor of the ratio of the two classes. The oversampling step was performed after stratification, to ensure that the same molecule is not repeated across training/validation/test sets.

\subsection{Network Design}
In this work, we used the Chemception CNN model without modification, and we refer readers to the original publication for architectural details.~\cite{goh2017c1} Briefly, Chemception is a CNN model that is optimized for handling sparse images of 2D representations of chemical structures, which incorporates the two major advances in recent CNN research: inception modules~\cite{szegedy2015} and residual learning.~\cite{he2015} A typical Chemception CNN model has alternating sequences of stacked Inception-Resnet blocks and Reduction blocks, and based on earlier work, we evaluated both the baseline T1\_F32 and optimized T3\_F16 architectures.~\cite{goh2017c1} In the nomenclature used, T\textit{x} refers to the general depth of the network (i.e. how many Inception-Resnet blocks were stacked), and F\textit{x} refers to the number of filters in the convolutional layers.

\subsection{Training the Neural Network}
Chemception was trained using a Tensorflow backend~\cite{abadi2016} with GPU acceleration using NVIDIA CuDNN libraries\cite{chetlur2014}. The network was created and executed using the Keras 1.2 functional API interface~\cite{chollet2015}. We use the RMSprop algorithm~\cite{hinton2012} to train for 100 epochs for each chemical task, using the standard settings recommended (learning rate = $10^{-3}$, $\rho = 0.9$, $\epsilon = 10^{-8}$). We used a batch size of 32, and also included an early stopping protocol to reduce overfitting. This was done by monitoring the loss of the validation set, and if there was no improvement in the validation loss after 25 epochs, the last best model was saved as the final model. In addition, during the training of Chemception for all image representations tested, we performed additional real-time data augmentation to the image using the ImageDataGenerator function in the Keras API, where each image was randomly rotated between 0 to 180 degrees.

For classification tasks (Tox21, HIV) we used the binary crossentropy loss function, and the evaluation metric reported in our paper that determines model’s performance is the area under the ROC-curve (AUC). For regression tasks (FreeSolv) we used the mean-squared-error loss function, and RMSE was the evaluation metric. Unless stated otherwise, the reported results in the paper are the mean values obtained from the 5-fold cross validation runs.

\section{Design of Image Representations}
\label{sec:design} 

In this section, we outline the design principles of augmented images in adding localized (i.e. pixel-specific) information to image channels, and reduced images in simplifying the information content encoded. 

\subsection{Design Principles}

In the original Chemception paper, single-channel greyscale images, hereon referred to as Standard images that depict molecular drawings was used.~\cite{goh2017c1} A number corresponding to the atomic number (i.e. position in the periodic table) was used as a unique identifier for different atoms, and chemical bonds were identified as "2". Apart from this data, no additional information was encoded in the image channels.

\begin{figure}[!htbp]
\centering
\includegraphics[scale=0.4]{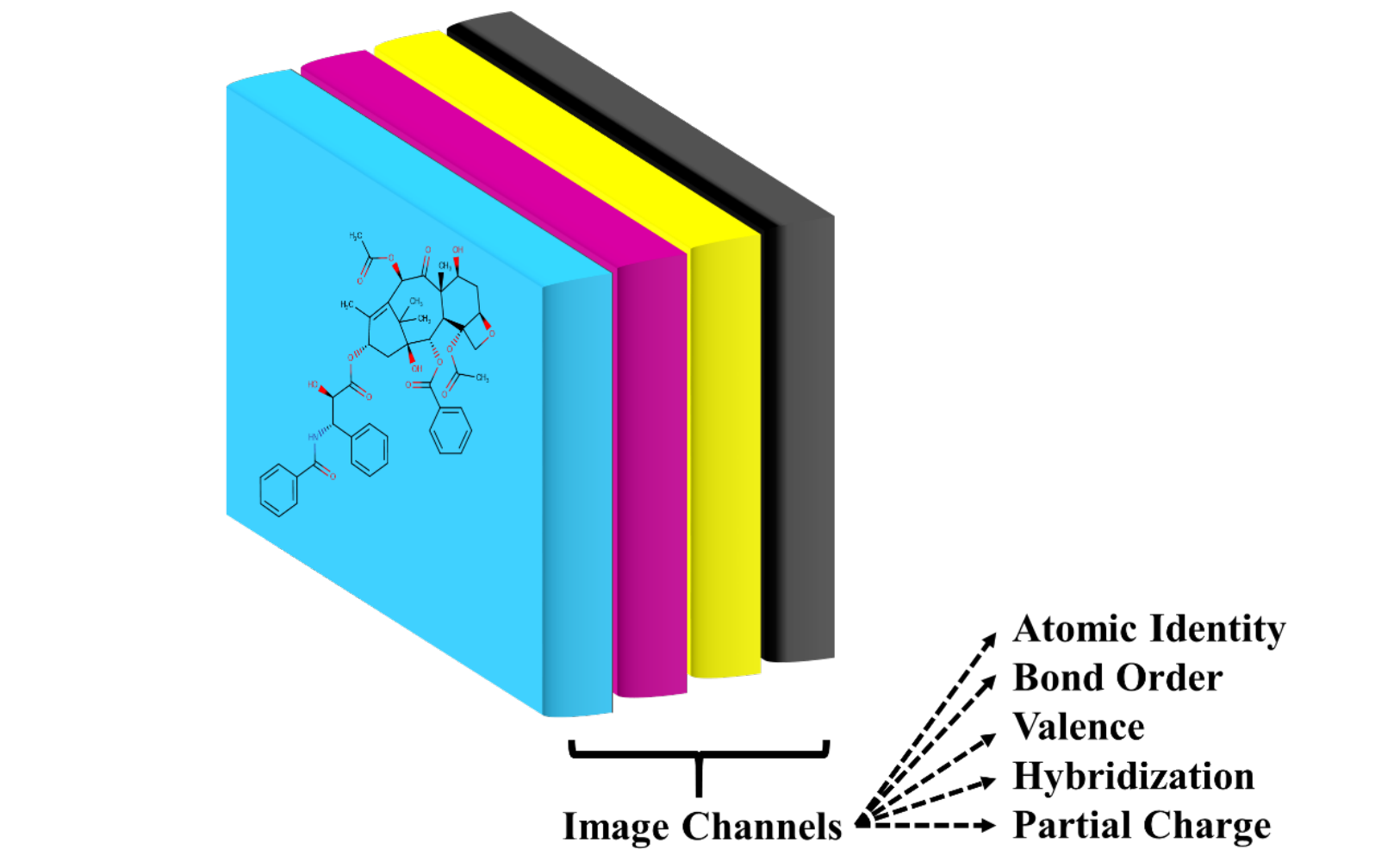}
\caption{\small Instead of encoding color information into the image channels, atom and bond-specific chemical information are encoded in the augmented image representation.}
\label{fig:2}
\end{figure}

In computer vision research, colored images are commonly used, where information relevant to color is encoded in multiple image channels, typically using a 3-channel RGB or 4-channel CMYK color model. Adopting this principle of encoding additional information into image channels, for our work, we have encoded localized (i.e. pixel-specific) atom and/or bond specific chemical information into a 4-channel image format to create a set of augmented image representation (Figure~\ref{fig:2}). We anticipate that the addition of appropriate localized chemical information would accelerate Chemception training, as this information can be used directly, and the neural network does not need to learn additional representations for information already provided in the image channels. We also hypothesize that the benefits of training on augmented images will be useful in the small labeled data limit, where there may be insufficient data for the neural network to develop optimal internal representations. One caveat is that any additional information encoded will only be useful, so as long as they are relevant to the task to be predicted – either through direct correlation or through the formation of higher-level representations that correlates with the task.

In this work, our goal is to develop a general-purpose format for encoding chemical images using the following design principles: (i) it should be quick to compute and scalable, and (ii) it should only encode basic chemical information. The rationale for the first principle is to ensure scalability to large chemical databases. In addition, to maintain compatibility with existing image analysis tools, we also limited the maximum number of channels to 4, which is analogous to 4-channel CMYK images. For the second design principle, we hypothesize that it would be more effective and generalizable to provide the neural network with the building blocks to construct more complex features, and thus we favor the selection of basic chemical information that is also incidentally inexpensive to compute.

\subsection{Schema of Augmented and Reduced Images}

Based on these design principles, we selected several computable properties for encoding, which includes: bond order, hybridization, valency, and partial charge. Technically, information pertaining to bond order can be inferred from the general topology of the 2D structure, but this representation requires sufficient data for deep neural networks to learn, and therefore there may still be benefits in encoding it directly. Hybridization in this context refers to the type of hybridization (sp, sp2, sp3, etc.) of the atom, and valence refers to the number of explicit connections an atom is bonded to. When combined together in the appropriate context, an atom's hybridization and valence can lead to inferences about bond order, lone pairs, and other related basic chemical concepts. Lastly, partial charge was computed by the methods developed by Gasteiger et. al,\cite{gasteiger1980} which represents one of the more basic electrostatics descriptions of an atom that can be computed quickly.

\begin{table}[!t] 
		\begin{center}
		\begin{tabular}{|c|c|c|}
				\hline
				Representation & Channels & Description \\
				\hline\hline
				Std & 1 & \makecell{Atomic no. + bond} \\
				\hline
				RedA & 1 & \makecell{"1" for atom} \\
				\hline
				RedB & 1 & \makecell{"1" for atom, "2" for bond} \\
				\hline
				EngA & 4 & \makecell{1. Atomic no. \\ 2. Bond order \\ 3. Partial charge \\ 4. Hybridization} \\
				\hline
				EngB & 4 & \makecell{1. Atomic no. \\ 2. Bond order \\ 3. Partial charge \\ 4. Valence} \\
				\hline
				EngC & 4 & \makecell{1. Atomic no. \\ 2. Bond order \\ 3. Valence \\ 4. Hybridization} \\
				\hline
				EngD & 4 & \makecell{1. Atomic no. + bond \\ 2. Partial charge \\ 3. Valence \\ 4. Hybridization} \\
				\hline
				Noise & 1 & \makecell{Random scattering of "1"} \\
				\hline
				Truth & 1 & \makecell{"1" for active, \\ only for atom/bond pixels} \\
				\hline
				Scrambled & 1 & \makecell{Unique random no. \\ for each atom/bond pixel} \\
				\hline
		\end{tabular} 
		\end{center}
\caption{Summary of the schema of various image representations evaluated.}
\label{table:2}
\end{table}

Using these basic computable properties, several permutations of image representations were evaluated. As summarized in (Table~\ref{table:2}), the schema for EngA, EngB, and EngC provides a different channel for atomic identity (encoded as atomic number) and bond identity (encoded as bond order). EngD compresses both atomic and bond identity information into a single channel (as per standard images used in Chemception), and adds all 3 atom-specific information into the remaining channels.

We also explored alternative single-channel image representations to infer the nature and type of chemical representations Chemception may have learned from its training data. For this approach, we start with standard images, and systematically reduced content from this channel. In the first level of simplification, RedB, all atoms are now identical and assigned "1", while bonds are assigned "2", which effectively makes the identity of different atoms indistinguishable. In RedA, bonds are removed from the images and only atoms remain. Furthermore, we devised a Noise representation, which is a random dispersion of noise values ("1"). This representation was intentionally constructed to be not-a-molecule, thus serving as a negative control to determine if Chemception was making predictions from noise. In a similar fashion, we constructed a Truth representation to serve as positive control. In this representation, we systematically assigned "1" for every atom and bond pixel if the molecule's property is active. Lastly, based on the results of our work, it also motivated the development of a Scrambled representation to test for the importance of chemical periodicity. In this representation, each atom and bond is assigned a random but unique number, ensuring that they can still be uniquely identified, but information about chemical periodicity which may be inferred from atomic number information is not available.

\section{Experiments}
\label{sec:exp} 

In this section, experiments were performed to determine the best augmented image representation. The resulting Augmented Chemception (AugChemception) CNN model was then compared against other contemporary DNN models in the literature. Lastly, by exploring the effectiveness of various augmented and reduced images, and using appropriate comparisons in the literature, we identified distinct learning patterns that suggests Chemception is learning about its tasks in the manner that is consistent with established chemical knowledge.

\subsection{Impact of Image Channel Information on Toxicity Prediction}

\begin{figure}[!htbp]
\centering
\includegraphics[scale=0.29]{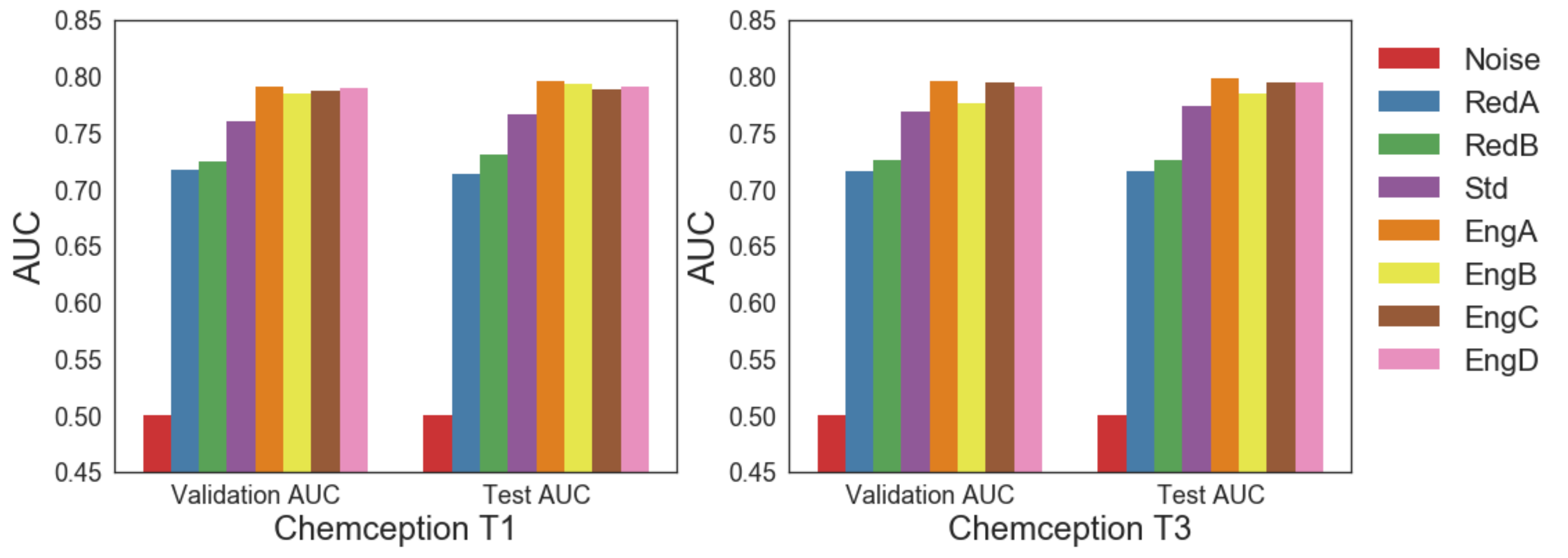}
\caption{\small Chemception performance on toxicity prediction when trained on various reduced and augmented images.}
\label{fig:3}
\end{figure}

We trained the baseline Chemception T1\_F32 model and optimized Chemception T3\_F16 model on the Tox21 dataset using both reduced and augmented images and the results are summarized in Figure ~\ref{fig:3}. The previous best model achieved a validation/test AUC of 0.768/0.773 using standard images. With augmented images, the best performance achieved was a validation/test AUC of 0.796/0.798 for the optimized Chemception model using the EngA image representation. We also observed that other engineered representation (EngB, EngC, EngD) achieved comparable performance, albeit in the 0.78 to 0.79 range. All augmented images were better than standard images which indicate that the additional information encoded was used by the model to improve its performance.

Next, we examine the performance of using reduced images. When trained on a set of RedA images, both Chemception T1\_F32 and Chemception T3\_F16 achieved similar validation/test AUC of 0.717/0.713 and 0.716/0.716 respectively. The fact that the AUC is not at 0.5 indicates the neural network has learned some chemistry to distinguish between toxic and non-toxic molecules just from the relative positions of generic indistinguishable atoms. As we add more chemical information to RedB images, where bonds are explicitly indicated, we observed a marginal improvement to a validation/test of AUC 0.724/0.731 and 0.726/0.726 for each respective model. These results suggest that explicit knowledge of bonds was apparently not the most important requirement for determining molecular toxicity. Given how fundamentally important the concept of chemical bonds is in chemistry, at first glance this observation is paradoxical. However, one has to recall that a bond is an artificial construct introduced to denote the linkages between various atoms, and one of its key role in chemistry research is to make it easier for chemists to formulate more sophisticated concepts starting with the notion of a bond.

Lastly, we note that the images used in training Chemception are extremely sparse compared to natural images used in typical computer vision research; this means that typically less than 10\% of the image has usable information (since the other 90\% is empty), and it is likely that representations learned by the neural network would have to identify relevant data in a sub-1\% portion of the image. Due to this peculiar image characteristic, a significant concern is that the neural network representations might not be robust, as a random scattering of pixels may not look that different from a valid molecular structure. To account for this possibility, we examine the performance of the control set of images. Training on noise images achieved a validation/test AUC of 0.5/0.5, which means that the resulting model has zero predictive power. At the opposite spectrum, training Chemception on the truth images resulted in a validation/test AUC of 1.0/1.0. Both sets of control experiments verify that Chemception is not learning "something from nothing", and also confirms the ability of the neural network to extract out relevant information, even from an extremely sparse image. 

\begin{figure}[!htbp]
\centering
\includegraphics[scale=0.29]{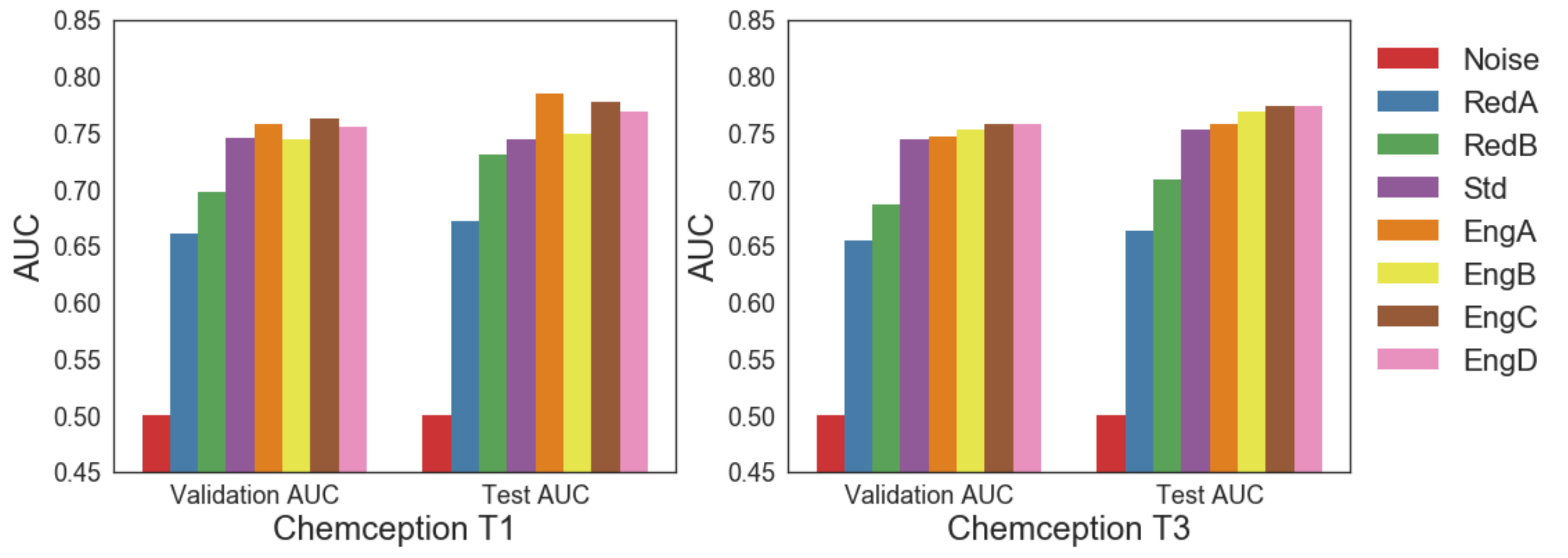}
\caption{\small Chemception performance on HIV activity prediction when trained on various reduced and augmented images.}
\label{fig:4}
\end{figure}

\subsection{Generalizing to Other Chemical Properties}

Having demonstrated that Chemception, when trained on augmented images, can lead to increased model performance, we now address the question of whether this effect can be generalized to other chemical properties. Here, we examine model performance on two different type of chemical properties that are not related to toxicity, namely HIV activity (a biochemical property) and solvation free energy (a physical property).

The results of the HIV activity prediction is summarized in Figure ~\ref{fig:4}. The previous best model achieved a validation/test AUC of 0.745/0.744 using standard images. The best performance achieved using augmented images was a validation/test AUC of 0.762/0.777 for the baseline model and 0.757/0.773 for the optimized model using the EngC image representation. Similar performance metrics were observed for other engineered representations (EngA, EngB, EngD) that achieved a comparable performance in the 0.75 range. In addition, the best representation for toxicity prediction (EngA) was not the best representation for activity prediction (EngC), although we observed that the difference in AUC metrics is minimal. As with toxicity, using augmented images provided consistently better results. Our results of reducing chemical information for activity prediction also parallels that observed for toxicity prediction, where increasing information content from RedA to RedB saw a gradual improvement from 0.660/0.672 to 0.698/0.730 for the baseline Chemception model, and a similar trend was observed for the optimized model. 

\begin{figure}[!htbp]
\centering
\includegraphics[scale=0.29]{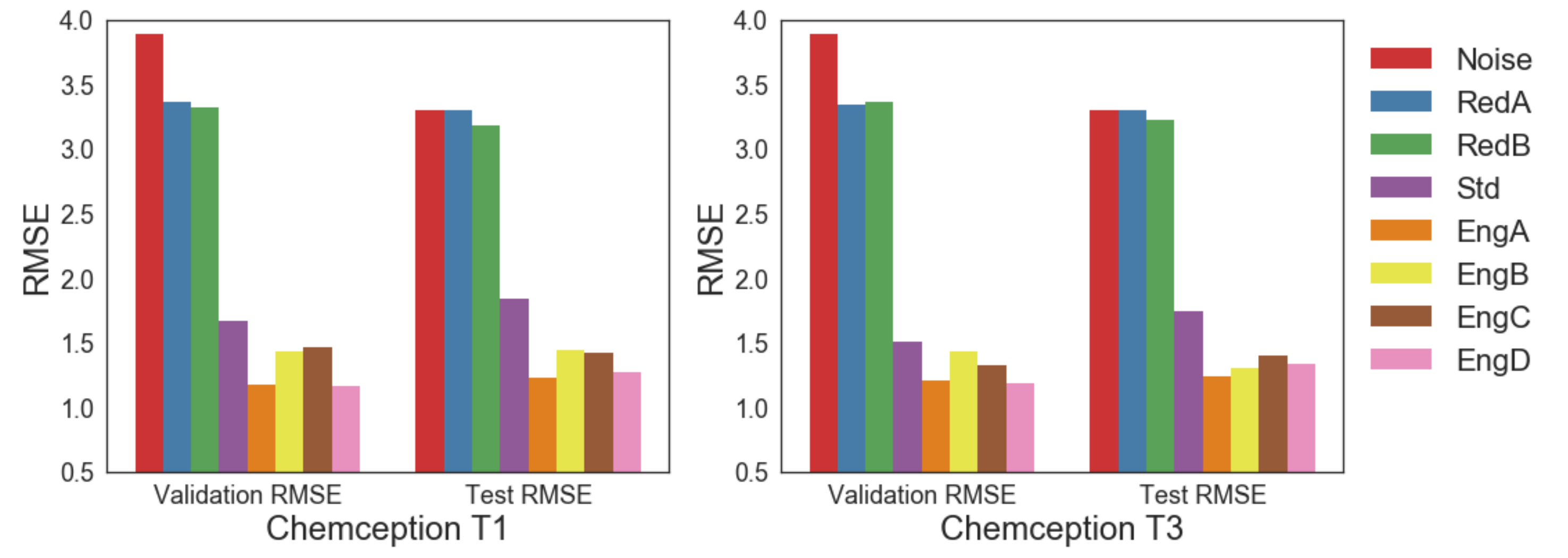}
\caption{\small Chemception performance on solvation free energy prediction when trained on various reduced and augmented images.}
\label{fig:5}
\end{figure}

The last chemical property is solvation free energy, which uses a much smaller dataset of 600 molecules. The previous work using standard images achieved a validation/test RMSE of 1.51/1.75 kcal/mol. As summarized in Figure ~\ref{fig:5}, augmented images substantially improved the model’s performance. The best performance achieved was a validation/test RMSE of 1.16/1.27 kcal/mol for the baseline Chemception model and 1.21/1.24 kcal/mol for the optimized Chemception model using the EngD image representation. We also observed that unlike toxicity/activity predictions where all 4 augmented image representations have performed comparably to one another, there is a difference in EngA and EngD accuracy (1.1-1.2 kcal/mol) relative to EngB and EngC (1.3-1.4 kcal/mol). Upon further examination, both EngA and EngD had partial charge and hybridization information encoded, but EngB and EngC do not. Using chemistry intuition, it is not unexpected that partial charge would be extremely relevant to predict solvation free energies as it serves as a proxy for the electrostatic nature of the atom and its local environment.

Next, we examine the performance of reduced images. The noise images attain a validation/test RMSE of 3.89/3.30 kcal/mol. From our earlier observations of Chemception models trained on noise images in toxicity and activity, this has resulted in a model of zero predictive power, and as such the above-specified RMSE value may be a reasonable zero baseline. Our results indicate that reduced images achieved RMSE metrics of 3.1-3.3 kcal/mol, which when compared to the 1.51/1.75 kcal/mol accuracy of standard images, are significantly worse and closer to the null results obtained from noise images. We note that this performance trend is unlike that observed for toxicity and activity predictions, where reduced images had accuracy that was more similar to standard images than noise images. 

\textit{To summarize, the collective results of various experiments thus far indicate that using the augmented image representations developed in this work has provided consistent performance improvement to Chemception's accuracy relative to standard images. In addition, this improvement appears to be consistent across different datasets and chemical properties, which suggests the performance lift obtained from using augmented images should be generalizable to other chemical properties.}

\subsection{AugChemception Performance Against Contemporary Models}

Having established that using augmented images provide a consistent performance improvement, we will now compare the performance of AugChemception against other contemporary DNN models reported in the literature. In particular, we will account for both performance metrics as well as the amount of chemical information that is used and/or provided as input data. The baseline model for comparison is the multilayer perceptron (MLP) deep neural network model (random data splitting) reported in the MoleculeNet benchmark.~\cite{wu2017}. In addition, we also compare to a novel convolutional graph algorithm (ConvGraph). Instead of providing explicit chemistry features, the ConvGraph algorithm uses an engineered graph representation to depict the molecule, where the nodes are annotated with localized chemical information and edges denote the connectivity between atoms.~\cite{kearnes2016} Conceptually, the ConvGraph algorithm and augmented images reported in this work are similar, as both methods use an engineered representation of the data as opposed to relying on explicit feature engineering. However, the key difference is that our method encodes the molecule in an image format, whereas ConvGraph encodes it as a graph. Based on this assessment, we order the level of chemical information provided in the following ascending order: reduced images, standard images, augmented images, molecular graphs, molecular fingerprints.

\begin{figure}[!htbp]
\centering
\includegraphics[scale=0.27]{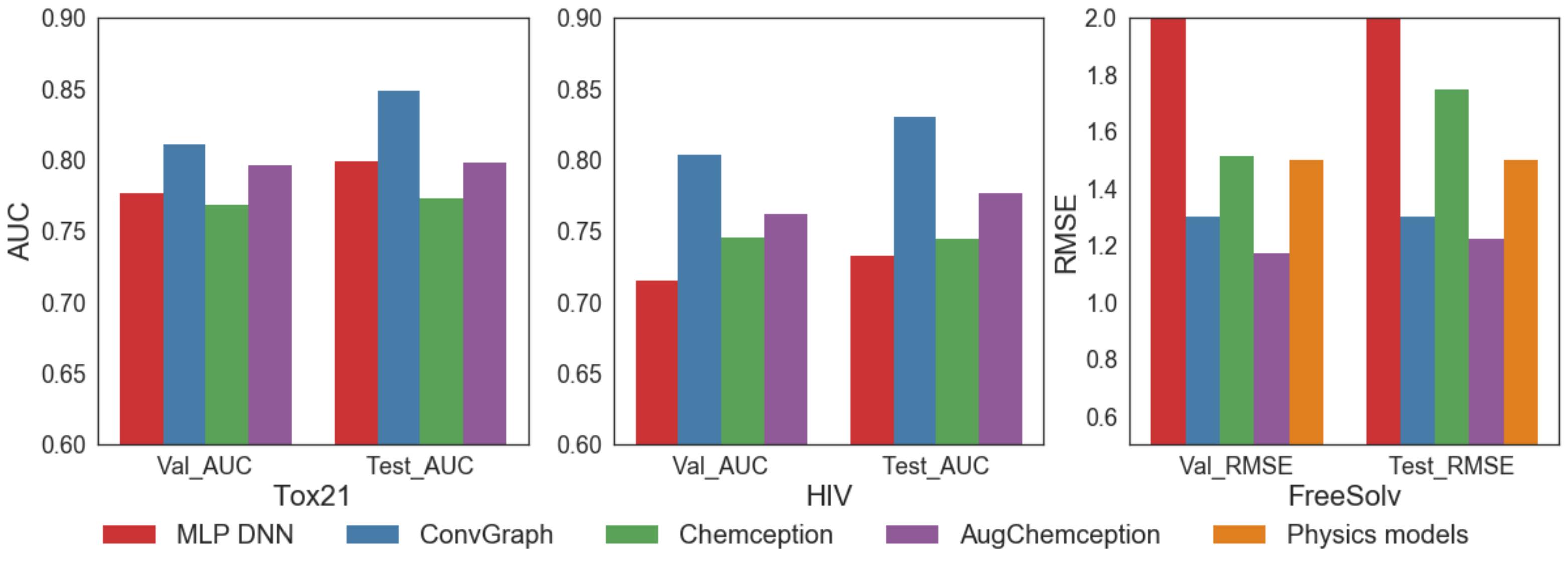}
\caption{\small AugChemception performance relative to contemporary deep learning models, and a novel convolutional graph algorithm.}
\label{fig:6}
\end{figure}

In Figure~\ref{fig:6}, we summarize the best AugChemception results relative to other DNN models listed above. For toxicity and activity predictions, AugChemception outperforms the baseline MLP model. This is despite the fact that the MLP model was trained with more chemical information, as it uses engineered features (molecular fingerprints) as input data, while AugChemception even with augmented images only benefits from 3 additional pieces of basic chemical information. AugChemception's better performance provides evidence that leveraging the representation learning ability of deep CNN models on raw data is a viable approach, and it may reduce the necessity of manual feature engineering of explicit chemistry features. We also note that when compared to ConvGraph, AugChemception trails in performance. In contrast, for solvation free energy prediction, AugChemception is the best performing model across all methods (RMSE 1.17/1.22 kcal/mol), outperforming both ConvGraph and physics-based models. At its current level of performance, AugChemception is approaching the "holy grail" of \textasciitilde1.0 kcal/mol chemical accuracy, and we acknowledge that similar levels of accuracy have been reported for other physical properties in recent DNN studies.~\cite{smith2017,schutt2017}.

\subsection{Inferring Chemception's Learning}

Deep learning is a "black box" algorithm, and as such it is challenging to directly determine how or what Chemception learns as it is trained to predict various chemical properties. However, given that we have systematically added and removed information into the image channels, and also using the results from comparable DNN models from the literature, it is possible to make inferences based on the consistent patterns in model performance, which may formulate additional hypothesis on what or how DNN models such as Chemception learns from chemical data. 

First, we summarize the key observations. For toxicity and activity prediction, training on reduced images provide less predictive accuracy, but still resulted in Chemception models that were fairly predictive of their respective tasks. In contrast, for solvation free energy prediction, using reduced images resulted in an accuracy that is close to that when trained on noise images. Therefore, based on the difference in model accuracy improvement with increasing chemical information, toxicity/activity vs. free energy of solvation predictions appear to follow two distinct learning patterns.

The current approach in toxicity research involves the identification of toxicophores (a specific pattern in how atoms are arranged) to predict molecular toxicity.~\cite{williams2002,muster2008} For HIV activity, the current structure-based research paradigm is premised on finding the appropriate functional groups (a common arrangement of a set of atoms) that can interact well with the target the molecule binds to.~\cite{mason2001,yang2010} From the observed trends in Chemception's accuracy on toxicity/activity modeling, it is evident that just the position of atoms is sufficient to construct a model that is more accurate than a null model. With reduced images, the only representations that can be reasonably learned are patterns of how atoms are arranged, which is analogous to identifying functional groups. Therefore, this suggests that Chemception is operating primarily as a functional group identifier, which is consistent with current understanding of toxicity and activity.

We now contrast this to solvation free energies, which unlike toxicity/activity prediction, is a physical property that can be computed from physics-based simulation methods. Reduced images are almost as inaccurate as a model trained on noise images, and a significant improvement in accuracy is noted when atomic number information is first introduced. This suggests that learning to be just a functional group identifier is not sufficient, and atomic number is a key piece of information. From the atomic number, it is possible for the neural network to form representations related to periodic trends, which is a key concept in chemistry, from which other properties like number of electrons can be inferred. This finding is relevant because contemporary physics-based simulation methods for calculating free energies incorporate some of the above-mentioned information. Therefore, we suggest that the significant improvement in Chemception's accuracy with the introduction of atomic number is an indicator that it is learning in a manner similar to established physics-based models.

\subsection{Validation of Chemception's Learning Patterns}

To validate this hypothesis on the importance of atomic number, we developed the scrambled image representation that contains almost the same amount of information as standard images. In this representation, atoms can be uniquely identified, but information about their periodicity is lost. Chemception models trained on scrambled images achieved a validation/test AUC of \textasciitilde0.74 for toxicity prediction and \textasciitilde0.72 on HIV activity prediction. Compared to the results of using standard images (AUC \textasciitilde0.77 for Tox21, \textasciitilde0.74 for HIV) relative to noise images (AUC 0.5 for Tox21 and HIV), scrambling the atomic identities had little effect, which is to be expected if the network is primarily learning to identify specific patterns of pixels in the image (i.e. functional groups). This result is in stark contrast to the performance of scrambled images on free energy of solvation prediction, which achieved a validation/test RMSE of \textasciitilde3.1 kcal/mol. Compared to the results of using standard images (RMSE \textasciitilde1.5 kcal/mol) relative to noise images (RMSE \textasciitilde3.9 kcal/mol), the huge gap in performance indicates that Chemception is using the atomic number information in its calculation of solvation free energies.

Next, we identify another trend by comparing Chemception to ConvGraph, where Chemception systematically underperforms in activity and toxicity prediction but outperforms in solvation free energy prediction. This is despite the fact that almost the same type of chemical information is being provided in either data representation. Our earlier hypothesis that Chemception operates as a functional group identifier for activity and toxicity prediction, explains the apparent underperformance relative to the ConvGraph algorithm. This is because when the molecule is represented as a graph, its topology is implicit to the data representation, making it easier to identify topological patterns, which will facilitate functional group identification. In contrast, for an image, specific representations will need to be developed by Chemception to understand molecular topology. Next, we observed that the ConvGraph algorithm encoded atomic identity as a one-hot vector, as opposed to our work that used atomic number directly. This means that periodicity will be difficult to infer. Our earlier hypothesis that atomic number enables Chemception to learn similarly to physics-based models, explains the apparent underperformance of the ConvGraph algorithm relative to Chemception.

\subsection{How Much Chemistry Does a Neural Network Need to Know?}

In this final section, we address the underlying question behind this work: \textit{How Much Chemistry Does a Deep Neural Network Need to Know to Make Accurate Predictions?} Based on our findings, only basic chemical information is necessary for a deep neural network to predict complex chemical properties, and advanced chemistry knowledge is evidently not a pre-requisite to achieve accurate predictions. In addition, we have identified that atomic number information is critical for predicting solvation free energies, and possibly also other physical properties that can be computed from physics-based  methods.  We also observed the high efficiency at which Chemception uses relevant information, as only less than 1\% of the input data was altered between various image representations.

\textit{The collective results therefore indicate that using deep neural networks can be a viable first approach to solve novel chemistry problems by serving as a baseline model. The corollary to this statement is: if a well-trained neural network provided with sufficient data cannot achieve the desired accuracy, it is likely that the data provided is not entirely relevant to the task being predicted.}

\subsection{Image Augmentation in Other Domains}

While the image augmentation techniques we have developed in this work is unique to chemistry, 3 key findings and design principles can be generalized to other domains. (i) In situations where there is limited labeled data, encoding relevant domain-specific information into image channels can be used to bootstrap the training of a deep neural network and increase overall model accuracy. (ii) Encoding localized pixel-specific information is recommended as it complements well with a CNN's ability to handle spatial correlations in images. (iii) Encoding basic information, which is analogous to providing conceptual building blocks to the neural network helps to generalize across different tasks in the same domain. \textit{Therefore, we anticipate that our image augmentation techniques will have considerable impact in fields that have substantial prior domain-specific research, which includes many scientific, engineering and financial modeling applications.}

\section{Conclusions}
\label{sec:conclusions}

In conclusion, we have developed a series of augmented images for use in training Chemception, a deep CNN model that predicts chemical properties using images of molecular diagrams. By encoding localized basic chemical information into the image channels, we have improved the accuracy of Chemception, where it now consistently outperforms contemporary MLP deep learning models trained on engineered chemistry features. Specifically, Chemception achieved a validation/test AUC of 0.796/0.798 for toxicity prediction, 0.762/0.777 for HIV activity prediction and a validation/test RMSE of 1.17/1.22 kcal/mol for solvation free energy prediction. Furthermore, by altering the information content encoded in the images, we identified two distinct learning patterns that appears to parallel established chemistry knowledge. Specifically, we infer that Chemception learns about toxicity and activity primarily as a functional group identifier, which is analogous to current research in identifying functional groups responsible for toxicity (i.e. toxicophores) or activity (i.e. pharmacophores). For solvation free energy, we show that atomic number information is necessary for Chemception to achieve predictive accuracy, and this importance of chemical periodicity parallels the underlying principles behind current physics-based simulations. Therefore, our findings indicate that Chemception will be a valuable tool in the future of deep learning assisted chemistry in a data-driven research paradigm, where it can be used as a first-pass approach in developing baseline models for novel research problems.

{\small
\bibliographystyle{ieee}
\bibliography{egbib}
}

\end{document}